\documentclass[letterpaper]{article} 
\usepackage{aaai2026}  
\usepackage{amsfonts}
\usepackage{times}  
\usepackage{helvet}  
\usepackage{courier}  
\usepackage[hyphens]{url}  
\usepackage{graphicx} 
\usepackage[numbers]{natbib}  
\usepackage{caption} 
\usepackage{amsmath}
\usepackage{graphicx}
\usepackage{subcaption}
\usepackage{amssymb}
\usepackage{booktabs}
\usepackage{multirow}
\usepackage{algorithm}
\usepackage{algorithmic}
\usepackage{newfloat}
\usepackage{listings}
\DeclareCaptionStyle{ruled}{labelfont=normalfont,labelsep=colon,strut=off} 
\floatstyle{ruled}
\newfloat{listing}{tb}{lst}{}
\floatname{listing}{Listing}
\title{Rethinking Selectivity in State Space Models: A Minimal Predictive Sufficiency Approach}
\author{
    Yiyi Wang\textsuperscript{\rm 1},
    Jian'an Zhang\textsuperscript{\rm 2, \rm 3},
    Hongyi Duan\textsuperscript{\rm 4, \rm 1}\thanks{Work done mainly while studying at XJTU.},
    Haoyang Liu\textsuperscript{\rm 2, \rm 5},
    Qingyang Li\textsuperscript{\rm 1}
}
\affiliations{
    \textsuperscript{\rm 1}Xi'an Jiaotong University\\
    \textsuperscript{\rm 2}Peking University\\
    \textsuperscript{\rm 3}Shanghai University of Finance and Economics\\
    \textsuperscript{\rm 4}The Hong Kong University of Science and Technology (Guangzhou)\\
    \textsuperscript{\rm 5}Xiamen University\\

}

\usepackage{bibentry}
\begin{document}

\maketitle

\begin{abstract}
State Space Models (SSMs), particularly recent selective variants like Mamba, have emerged as a leading architecture for sequence modeling, challenging the dominance of Transformers. However, the success of these state-of-the-art models largely relies on heuristically designed selective mechanisms, which lack a rigorous first-principle derivation. This theoretical gap raises questions about their optimality and robustness against spurious correlations. To address this, we introduce the Principle of Predictive Sufficiency, a novel information-theoretic criterion stipulating that an ideal hidden state should be a minimal sufficient statistic of the past for predicting the future. Based on this principle, we propose the Minimal Predictive Sufficiency State Space Model (MPS-SSM), a new framework where the selective mechanism is guided by optimizing an objective function derived from our principle. This approach encourages the model to maximally compress historical information without losing predictive power, thereby learning to ignore non-causal noise and spurious patterns. Extensive experiments on a wide range of benchmark datasets demonstrate that MPS-SSM not only achieves state-of-the-art performance, significantly outperforming existing models in long-term forecasting and noisy scenarios, but also exhibits superior robustness. Furthermore, we show that the MPS principle can be extended as a general regularization framework to enhance other popular architectures, highlighting its broad potential.
\end{abstract}


\section{1. Introduction}

\subsection{1.1 Background}
The evolution of deep learning for sequence modeling has been marked by a paradigm shift. Initially dominated by Recurrent Neural Networks (RNNs) and their gated variants like LSTMs, which suffer from gradient vanishing and difficulties in parallelization \cite{gu2024survey_s4}, the field later embraced Transformers for their prowess in capturing global dependencies via self-attention \cite{vaswani2017attention}. However, the subsequent "Transformer debate," sparked by the seminal work of \cite{zeng2023are}, challenged the "more complexity is better" mantra. It revealed a fundamental conflict between the permutation-invariant nature of self-attention and the inherent temporal order of time series data. This critique has catalyzed an "architectural renaissance," spurring exploration into diverse architectures with stronger inductive biases. Among these, State Space Models (SSMs), especially Mamba \cite{gu2023mamba}, have risen to the forefront, distinguished by their linear complexity and powerful long-range dependency modeling capabilities.

\subsection{1.2 The Core Problem}
While Mamba's empirical success, driven by its input-dependent selective mechanism, is undeniable, its theoretical foundation remains soft. The mechanism, which dictates how the state matrices $\mathbf{B}$, $\mathbf{C}$, and step size $\Delta$ adapt to inputs, is a powerful but largely heuristic innovation \cite{gu2023mamba, gu2024mamba_explained}. It lacks the first-principle derivation that grounds models like S4 \cite{gu2022efficiently} in the HiPPO theory of online function approximation \cite{gu2020hippo}. This theoretical void leaves a critical question unanswered: Can we derive a non-heuristic, theoretically guaranteed selective mechanism for SSMs from a more fundamental principle of optimal information processing?

\subsection{1.3 Our Contribution}
This paper introduces the Minimal Predictive Sufficiency State Space Model (MPS-SSM) to fill this theoretical gap. Our contributions are:
\begin{itemize}
    \item We propose a new core theoretical principle, the Principle of Predictive Sufficiency, which provides a first-principle information-theoretic guide for designing selective mechanisms. It posits that an ideal state should be a minimal sufficient statistic of the past for the future.
    \item Inspired by the goals of the Disentangled Information Bottleneck (DisenIB) \cite{pan2021disenib}, we formulate an objective that seeks maximal compression without sacrificing predictive power, avoiding the trade-offs inherent in the classic Information Bottleneck \cite{tishby1999information}.
    \item We design a novel, end-to-end trainable SSM architecture based on this theory, which learns to filter non-predictive information, thereby enhancing robustness.
    \item Through extensive experiments, we validate the state-of-the-art performance and superior robustness of MPS-SSM, and demonstrate the universality of our principle as a general regularization framework.
\end{itemize}

\subsection{1.4 Paper Structure}
Section 2 reviews related work. Section 3 details the theoretical derivation of MPS-SSM. Section 4 presents the experimental design and results. Section 5 concludes the paper.


\section{2. Related Work}
\subsection{2.1 Evolution of State Space Models (SSMs)}
The lineage of modern SSMs traces back to the HiPPO theory \cite{gu2020hippo}, which formalizes the optimal online projection of a function's history into a compressed state using orthogonal polynomials. This provided the theoretical foundation for the Structured State Space Sequence (S4) model \cite{gu2022efficiently}, which introduced a Normal Plus Low-Rank (NPLR) parameterization to make SSMs computationally feasible. S4 achieved near-linear time complexity by leveraging a dual recurrent-convolutional representation, establishing SOTA performance on long-range dependency benchmarks like the Long Range Arena (LRA) \cite{tay2020long}. However, S4 is a Linear Time-Invariant (LTI) system, limiting its ability to perform content-based reasoning. The Mamba architecture \cite{gu2023mamba} addressed this by introducing an input-dependent selective mechanism, making key parameters ($\mathbf{B}$, $\mathbf{C}$, $\Delta$) functions of the input. This selectivity, combined with a hardware-aware parallel scan algorithm, allows Mamba to achieve linear-time scaling and SOTA performance on information-dense modalities like language \cite{gu2024mamba_explained}. While empirically powerful, this selective mechanism is heuristic, motivating our search for a more principled, information-theoretic foundation.

\subsection{2.2 Information Theory in Representation Learning}
The Information Bottleneck (IB) principle \cite{tishby1999information} is a foundational framework for learning useful representations. It formalizes the trade-off between compressing an input $X$ into a representation $T$ (minimizing $I(X;T)$) and preserving predictive information about a target $Y$ (maximizing $I(T;Y)$). For deep learning, the Deep Variational Information Bottleneck (VIB) provides a practical implementation but faces optimization challenges like posterior collapse \cite{alemi2017deep}. Critically, the standard IB Lagrangian imposes a strict trade-off where any compression necessarily degrades predictive performance \cite{pan2021disenib}. The Disentangled Information Bottleneck (DisenIB) \cite{pan2021disenib} proposed an alternative objective that aims for maximal compression without sacrificing predictive power, seeking to learn a minimal sufficient statistic \cite{pan2021disenib}. Our work is inspired by this latter goal, proposing the Principle of Predictive Sufficiency as a more direct and stringent criterion specifically for sequential modeling. This aligns with the concept of "predictive information" \cite{bialek2001predictability}, where the state $h_t$ should be a minimal sufficient statistic of the past $X_{<t}$ for the future $X_{\geq t}$.

\subsection{2.3 Robustness and Generalization in TSF}
A central challenge in time series forecasting (TSF) is non-stationarity, where data distributions shift over time \cite{deng2024domain, jin2025battling}. This can cause models to learn spurious correlations that fail to generalize. The work of \cite{zeng2023are} highlighted this issue, showing that a simple linear model could outperform complex Transformers on several benchmarks, suggesting the Transformers were overfitting to non-robust features. This underscores the need for models with strong inductive biases that can distinguish predictive signals from noise. While defense mechanisms like normalization \cite{kim2021reversible} and test-time adaptation \cite{jin2025battling} have been proposed, our MPS-SSM directly addresses this challenge at the architectural level by formalizing the objective of filtering out non-predictive information, which, as our theory and experiments show, leads to enhanced robustness.


\section{3. Methodology}

This chapter lays the theoretical groundwork for our new paradigm, the Minimal Predictive Sufficiency State Space Model (MPS-SSM). Moving beyond the direct application of existing information-theoretic tools to SSMs, we start from the fundamental goal of time series forecasting to propose a new core principle: The Principle of Predictive Sufficiency. We will show that this principle not only provides a more profound theoretical guide for designing sequential models but also naturally leads to guarantees of robustness against irrelevant noise and spurious correlations. Finally, we detail how this principle is translated into an operational objective function and an end-to-end trainable SSM architecture.

\subsection{3.1. Preliminaries: SSM}

The theoretical foundation of SSMs originates from the discretization of continuous-time systems to process sequential data.

\textbf{Definition 3.1 (Continuous-Time SSM):} A linear, time-invariant (LTI) SSM is described by the following ordinary differential equation (ODE):
\begin{align}
h'(t) &= \mathbf{A}h(t) + \mathbf{B}u(t) \\
y(t) &= \mathbf{C}h(t) + \mathbf{D}u(t)
\end{align}
where $u(t) \in \mathbb{R}^{M}$ is the input signal, $h(t) \in \mathbb{R}^{D}$ is the hidden state, and $y(t) \in \mathbb{R}^{N}$ is the output. The matrices $\mathbf{A} \in \mathbb{R}^{D \times D}$, $\mathbf{B} \in \mathbb{R}^{D \times M}$, and $\mathbf{C} \in \mathbb{R}^{N \times D}$ are the state, input, and output matrices, respectively.

To apply this to discrete time series data $u_k$, the continuous system is discretized using a fixed sampling period $\Delta$. A common method is the zero-order hold (ZOH), which yields the discrete-time SSM:

\textbf{Definition 3.2 (Discrete-Time SSM):}
\begin{align}
h_k &= \bar{\mathbf{A}}h_{k-1} + \bar{\mathbf{B}}u_k \\
y_k &= \mathbf{C}h_k + \mathbf{D}u_k
\end{align}
where $\bar{\mathbf{A}} = \exp(\Delta \mathbf{A})$ and $\bar{\mathbf{B}} = (\exp(\Delta \mathbf{A}) - \mathbf{I})\mathbf{A}^{-1}\mathbf{B}$.

The core innovation of modern SSMs like Mamba is the introduction of a selective mechanism, which makes the parameters $\mathbf{B}$, $\mathbf{C}$, and the step size $\Delta$ functions of the current input $u_k$:
$$ \Delta_k = G_\Delta(u_k), \quad \mathbf{B}_k = G_B(u_k), \quad \mathbf{C}_k = G_C(u_k) $$
where $G$ is a function parameterized by a neural network. This design endows the model with the ability to dynamically adjust its behavior based on content, but the functional form of $G$ is heuristic and lacks guidance from first principles.

\subsection{3.2. Core Theory}

To address the aforementioned problem from first principles, we propose the following core theoretical principle.

\textbf{Definition 3.3 (The Principle of Predictive Sufficiency):} For a time series process, the ideal hidden state $h_k$ at time $k$ should be the Minimal Predictive Sufficient Statistic of the observed history $U_{1:k}=\{u_1, ..., u_k\}$ for the future $Y_{k:\tau}=\{y_{k+1}, ..., y_{k+\tau}\}$. This must satisfy two conditions simultaneously:

\begin{enumerate}
    \item \textbf{Predictive Sufficiency:} The state $h_k$ must capture \textbf{all} predictive information contained in the history $U_{1:k}$ about the future $Y_{k:\tau}$. Formally, the future is conditionally independent of the past given the present state:
    $$ Y_{k:\tau} \perp U_{1:k} | h_k $$
    This is equivalent to the mutual information equality: $I(U_{1:k}; Y_{k:\tau}) = I(h_k; Y_{k:\tau})$.

    \item \textbf{Minimality:} Among all statistics that satisfy condition (1), $h_k$ must be the most concise, i.e., it contains the minimum possible amount of information about the history $U_{1:k}$. Formally, for any other sufficient statistic $h'_k$, $h_k$ is a deterministic function of it, which implies information-theoretic minimality:
    $$ I(U_{1:k}; h_k) \le I(U_{1:k}; h'_k) $$
\end{enumerate}

This principle is fundamentally different from the standard Information Bottleneck (IB). IB seeks a trade-off between compressing the past and predicting the future. In contrast, the Principle of Predictive Sufficiency allows for no trade-off. It demands the most extreme compression of the past, under the strict constraint that no predictive capability is lost. This is a stronger and more appropriate requirement for the task of prediction.

\subsection{3.3. Core Method: MPS-SSM}

We translate the Principle of Predictive Sufficiency into an operational optimization objective and model architecture.

\subsubsection{3.3.1. The Objective Function of MPS-SSM.}
According to Definition 3.3, our goal is to find a state update dynamic that satisfies sufficiency while minimizing the information complexity required by minimality, $I(U_{1:k}; h_k)$.

\begin{itemize}
    \item The Sufficiency condition is approximated via a standard prediction loss, $\mathcal{L}_{\text{Pred}}$. A model that minimizes prediction error must implicitly learn a hidden state that is predictively sufficient for the future.
    \item The Minimality condition is enforced through an explicit regularization term, $\mathcal{L}_{\text{Min}}$, which directly penalizes the mutual information between the hidden state and the history.
\end{itemize}

Thus, the total objective function for MPS-SSM is:
$$ \mathcal{L}_{\text{Total}} = \mathcal{L}_{\text{Pred}} + \lambda \cdot \mathcal{L}_{\text{Min}}(h; U) $$
where:
\begin{itemize}
    \item \textbf{Prediction Loss ($\mathcal{L}_{\text{Pred}}$):}
    $$
    \mathcal{L}_{\text{Pred}} = \frac{1}{T \cdot \tau} \sum_{k=1}^{T} \sum_{i=1}^{\tau} \mathbb{E}_{p(h_k|U_{1:k})} [||\hat{y}_{k+i}(h_k) - y_{k+i}||^2]
    $$
    \item \textbf{Minimality Regularizer ($\mathcal{L}_{\text{Min}}$):}
    $$
    \mathcal{L}_{\text{Min}} = \frac{1}{T} \sum_{k=1}^{T} I(U_{1:k}; h_k)
    $$
\end{itemize}
As direct computation and optimization of $I(U_{1:k}; h_k)$ is intractable, we employ a variational approximation, estimating its upper bound using an auxiliary decoder $p_\theta(u_k|h_k)$.

\subsubsection{3.3.2. Architecture and Training of MPS-SSM.}
The MPS-SSM architecture comprises:
\begin{enumerate}
    \item \textbf{SSM Backbone:} A standard SSM structure (e.g., based on Mamba) serves as the core.
    \item \textbf{Selection Gate Network $G_\phi(u_k)$:} Generates time-varying parameters $\{\Delta_k, \mathbf{B}_k, \mathbf{C}_k\}$ based on the current input $u_k$.
    \item \textbf{Minimality Regularization Module:} An auxiliary decoder $p_\theta(u_k|h_k)$ used to estimate and minimize $I(U_{1:k}; h_k)$ during training.
    \item \textbf{Prediction Head:} Maps the hidden state $h_k$ to the forecast $\hat{y}_k$.
\end{enumerate}
The entire model is trained end-to-end by minimizing the total objective function $\mathcal{L}_{\text{Total}}$.

\subsection{3.4. Theoretical Guarantees}
Our principle-based approach allows us to establish the following fundamental and powerful theoretical guarantees for MPS-SSM.

\textbf{Theorem 1 (Properties of the Optimal MPS-SSM Solution).}
Let the MPS-SSM model define a hypothesis space $\mathcal{H}$ for the hidden state dynamics $h$. Let its total objective be $\mathcal{L}_{\text{Total}}(h) = \mathcal{L}_{\text{Pred}}(h) + \lambda \cdot \hat{I}(U; h)$, where $\hat{I}(U; h)$ is a tight, data-dependent variational upper bound on the mutual information $I(U_{1:k}; h_k)$. Let $h^* \in \mathcal{H}$ be a global minimizer of $\mathcal{L}_{\text{Total}}(h)$.
\textbf{Assumption:} The model's function class is sufficiently expressive such that a solution exists for which $\mathcal{L}_{\text{Pred}}$ can approach zero.
\textbf{Conclusion:} The hidden state sequence $h^*_k$ defined by the global optimal solution $h^*$ converges in an information-theoretic sense to the minimal predictive sufficient statistic of the history $U_{1:k}$ for the future $Y_{k:\tau}$. Specifically, $h^*$ exhibits:
\begin{enumerate}
    \item \textbf{Sufficiency:} As $\mathcal{L}_{\text{Pred}}(h^*) \to 0$, the hidden state $h^*_k$ retains all predictive information from the history about the future, i.e., $I(h^*_k; Y_{k:\tau}) \to I(U_{1:k}; Y_{k:\tau})$.
    \item \textbf{Minimality:} Among all solutions that satisfy the approximate sufficiency, $h^*$ is the one that minimizes the information complexity between the history and the state, as measured by $\hat{I}(U; h)$.
\end{enumerate}
\textit{(Proof provided in Appendix D.1)}

\textbf{Theorem 2 (Theoretical Invariance to Non-Causal Perturbations).}
Let the input sequence be $u_k = u^{\text{sig}}_k + \epsilon_k$, where $u^{\text{sig}}_k$ is the true causal signal and $\epsilon_k$ is a non-causal perturbation, meaning the perturbation process $\{\epsilon_j\}_{j=1}^k$ is conditionally independent of the future $Y_{k:\tau}$ given the true history $U^{\text{sig}}_{1:k}$, i.e., $\{\epsilon_j\}_{j=1}^k \perp Y_{k:\tau} | U^{\text{sig}}_{1:k}$. Let $h^*$ be the global optimal solution described in Theorem 1.
\textbf{Conclusion:} Under the assumption of Theorem 1, the conditional probability distribution of the hidden state $h^*_k$ is approximately invariant to the non-causal perturbation $\epsilon_{1:k}$. Formally, the KL divergence between the two conditional distributions approaches zero:
$$ D_{KL}\left( p(h^*_k | U^{\text{sig}}_{1:k} + \epsilon_{1:k}) \parallel p(h^*_k | U^{\text{sig}}_{1:k}) \right) \to 0 $$
\textit{(Proof provided in Appendix D.2)}


\section{4. Experiments}

This section is dedicated to empirically validating our proposed MPS-SSM. We first detail the experimental setup. Then, we conduct an in-depth analysis of the core regularization parameter, $\lambda$, to demonstrate its role in balancing performance and robustness, providing empirical evidence for our theoretical claims.

\subsection{4.1 Experimental Setup}

\paragraph{Datasets.}
We conduct experiments on a comprehensive set of widely-used time series forecasting benchmarks: ETT (ETTh1, ETTh2, ETTm1, ETTm2), Weather, Electricity, Traffic, and Exchange. These datasets cover diverse domains and characteristics, allowing for a thorough evaluation of model performance and generalization.

\paragraph{Baselines.}
Our model is compared against a wide array of state-of-the-art (SOTA) models, including Transformer-based architectures (PatchTST, iTransformer, Autoformer, Informer, FEDformer), modern SSMs (Mamba), and strong linear models (DLinear, RLinear).

\paragraph{Evaluation Metrics.}
Following standard protocols, we use Mean Squared Error (MSE) and Mean Absolute Error (MAE) as the primary metrics for evaluation.

\paragraph{Implementation Details.}
All experiments were conducted on a server equipped with 8 NVIDIA RTX 4090 GPUs, running GNU/Linux (Kernel 5.15.0) and CUDA 12.4. Our MPS-SSM consists of an embedding layer, followed by a stack of MPS-SSM blocks (N=1 for simplicity in our main analysis), and a final prediction head. The key component is the minimality regularization module, which is implemented as a lightweight auxiliary decoder. To investigate the impact of the regularization strength, we test a wide range of values for $\lambda \in \{0, 0.01, 0.1, 1, 2, 5, 10, 100, 1000, 10^9\}$. A value of $\lambda=0$ represents the baseline without regularization, while a very large $\lambda$ simulates the effect of extreme information compression.

\begin{figure*}[t!]
    \centering
    \begin{subfigure}[b]{0.48\textwidth}
        \centering
        \includegraphics[width=\linewidth]{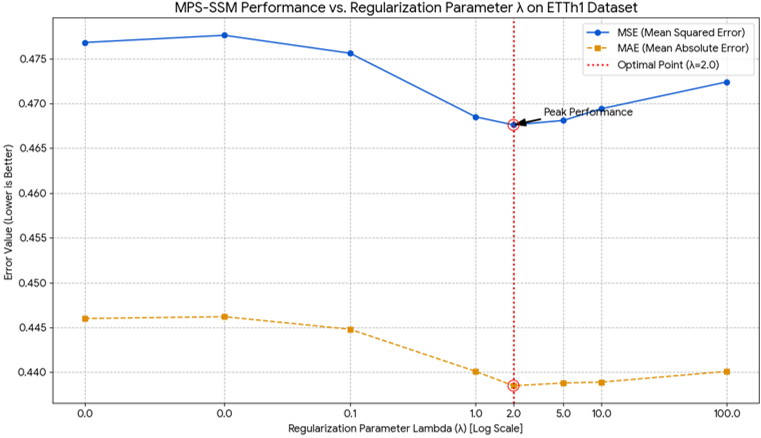}
        \caption{Performance vs. $\lambda$ on ETTh1 Dataset. The optimal performance is found at $\lambda=2.0$.}
        \label{fig:etth1_lambda}
    \end{subfigure}
    \hfill 
    \begin{subfigure}[b]{0.48\textwidth}
        \centering
        \includegraphics[width=\linewidth]{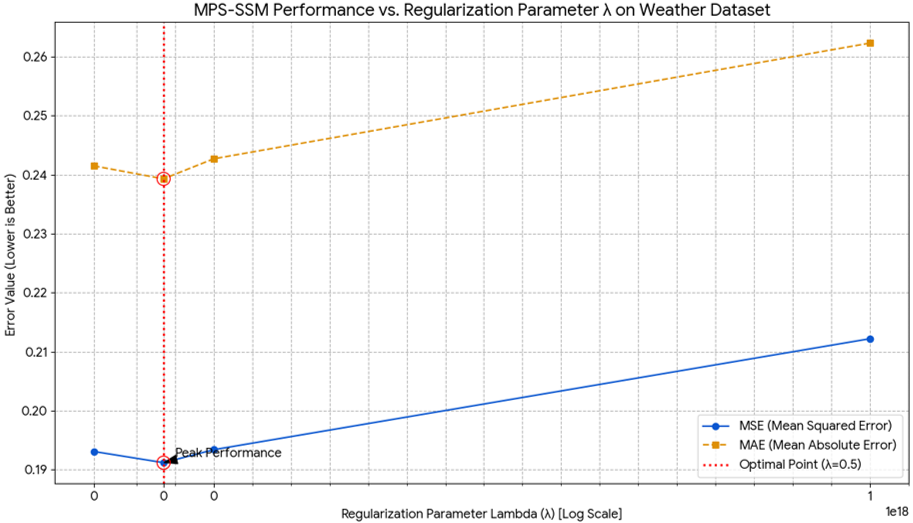}
        \caption{Performance vs. $\lambda$ on Weather Dataset. The optimal point is at a low $\lambda=0.5$.}
        \label{fig:weather_lambda}
    \end{subfigure}
    
    \vspace{1em} 

    \begin{subfigure}[b]{0.48\textwidth}
        \centering
        \includegraphics[width=\linewidth]{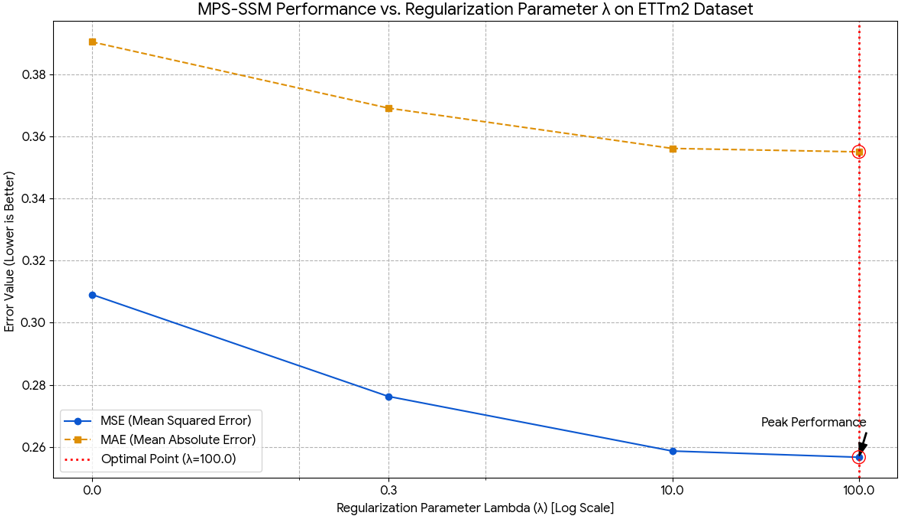}
        \caption{Performance vs. $\lambda$ on ETTm2 Dataset. A high regularization strength of $\lambda=100.0$ is needed.}
        \label{fig:ettm2_lambda}
    \end{subfigure}
    \hfill
    \begin{subfigure}[b]{0.48\textwidth}
        \centering
        \includegraphics[width=\linewidth]{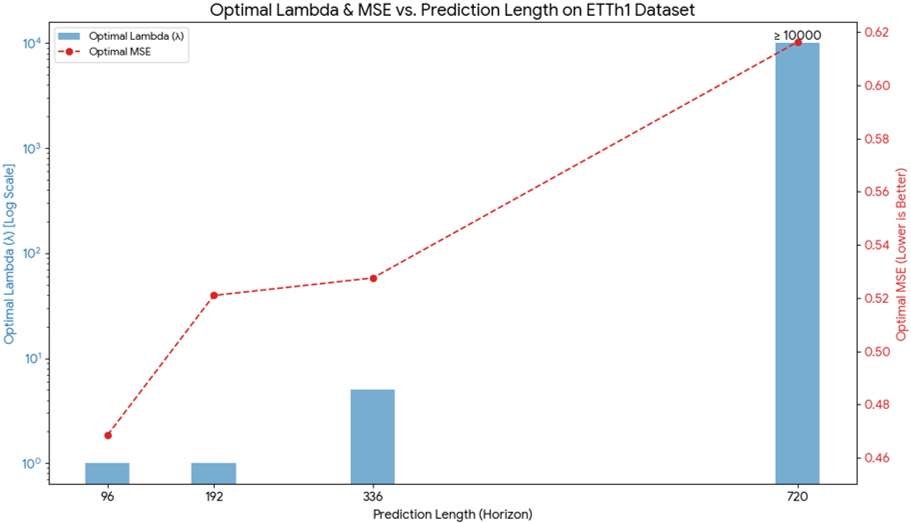}
        \caption{Optimal $\lambda$ and MSE vs. Prediction Length on ETTh1. Required $\lambda$ grows with the horizon.}
        \label{fig:lambda_vs_horizon}
    \end{subfigure}

    \vspace{1em}

    \begin{subfigure}[b]{0.48\textwidth}
        \centering
        \includegraphics[width=\linewidth]{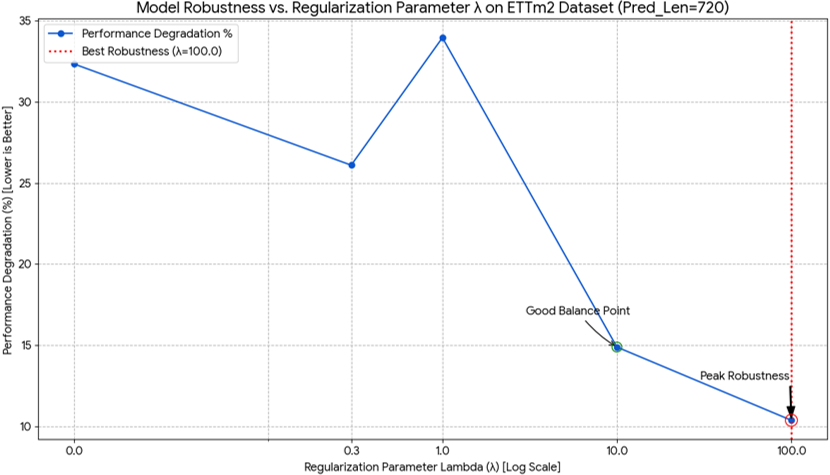}
        \caption{Model Robustness vs. $\lambda$ on ETTm2. Degradation decreases as $\lambda$ increases.}
        \label{fig:robustness_ettm2}
    \end{subfigure}
    \hfill
    \begin{subfigure}[b]{0.48\textwidth}
        \centering
        \includegraphics[width=\linewidth]{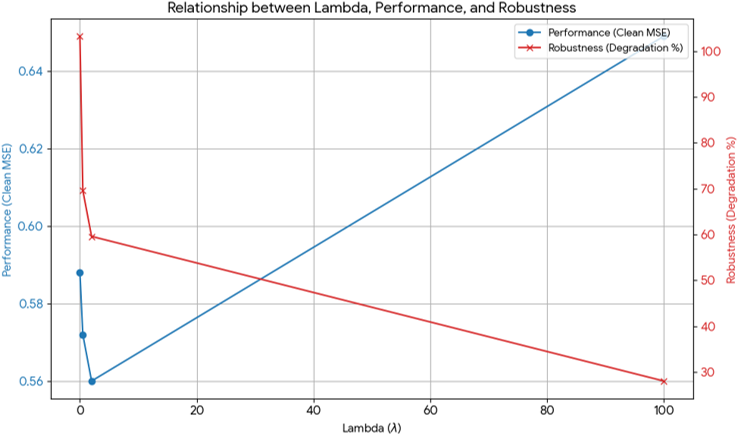}
        \caption{The trade-off between Performance (Clean MSE) and Robustness (Degradation \%) on ETTm1.}
        \label{fig:tradeoff_ettm1}
    \end{subfigure}

    \caption{Comprehensive analysis of the regularization parameter $\lambda$. (a-c) The "sweet spot" effect on performance across datasets with varying regularization needs. (d) The relationship between prediction horizon and optimal $\lambda$. (e-f) The impact of $\lambda$ on model robustness and the performance-robustness trade-off.}
    \label{fig:lambda_analysis_combined}
\end{figure*}

\subsection{4.2 Analysis of the Regularization Coefficient $\lambda$}

According to our Principle of Predictive Sufficiency, the regularization term controlled by $\lambda$ is crucial for encouraging the model to discard non-predictive information (noise), thereby enhancing generalization. This section empirically investigates the role of $\lambda$.

\subsubsection{4.2.1 The "Sweet Spot" Effect on Performance.}

We first analyze the performance trajectory of MPS-SSM as $\lambda$ varies. Our findings reveal a consistent "sweet spot" effect, where performance follows a U-shaped curve, indicating an optimal balance between retaining useful information and forgetting noise.

\paragraph{Case 1: Medium Regularization Demand (ETTh1).}
As shown in Figure \ref{fig:etth1_lambda}, on the ETTh1 dataset, the model's performance (both MSE and MAE) first improves as $\lambda$ increases from 0, reaches its peak at $\lambda=2.0$, and then degrades with larger $\lambda$ values. This perfectly illustrates the sweet spot phenomenon: a moderate degree of principled forgetting is beneficial, but excessive compression harms predictive capability by discarding useful signals.

\paragraph{Case 2: Low vs. High Regularization Demand.}
The optimal $\lambda$ is highly context-dependent, acting as a probe into the dataset's intrinsic properties. For the relatively clean Weather dataset (Figure \ref{fig:weather_lambda}), which has a high signal-to-noise ratio, a very small $\lambda=0.5$ is sufficient to achieve optimal performance. In contrast, for the more complex ETTm2 dataset (Figure \ref{fig:ettm2_lambda}), a much stronger regularization with $\lambda=100.0$ is required to force the model to learn robust long-term patterns. This demonstrates that the optimal regularization strength adapts to the data's complexity.

\paragraph{Adaptation to Task Difficulty.}
A profound finding is that the optimal $\lambda$ correlates with the difficulty of the prediction task. As illustrated in Figure \ref{fig:lambda_vs_horizon}, for the ETTh1 dataset, the optimal $\lambda$ grows exponentially as the prediction horizon increases from 96 to 720. To forecast the distant future, the model must ignore short-term fluctuations and focus on macro-level trends, which requires a stronger forgetting mechanism (a larger $\lambda$). This strongly suggests that MPS-SSM can adapt its information compression strategy to the task's demands, a key feature of our principled approach.

\subsubsection{4.2.2 Impact on Robustness: Empirical Validation of Theorem 2.}

We now empirically validate Theorem 2, which posits that the optimal state learned by MPS-SSM should be invariant to non-causal perturbations. We inject impulse noise into the test sets and measure the performance degradation.

\paragraph{Robustness Enhancement.}
As shown in Figure \ref{fig:robustness_ettm2}, for the ETTm2 long-forecasting task, the model's robustness (lower degradation percentage is better) monotonically improves as $\lambda$ increases. At $\lambda=100.0$, the model is not only the best performer on clean data but also the most robust, with its performance degradation reduced by nearly 3x compared to the unregularized ($\lambda=0$) baseline. This provides direct empirical evidence for Theorem 2: by penalizing information complexity, we successfully guide the model to ignore non-causal noise.

\paragraph{The Performance-Robustness Trade-off.}
Figure \ref{fig:tradeoff_ettm1} reveals a fascinating trade-off on the ETTm1 dataset. While robustness (red line) consistently improves with a larger $\lambda$, the performance on clean data (blue line) follows the U-shaped "sweet spot" curve. This means a user can make a deliberate choice: select a moderate $\lambda$ (e.g., 2.0) for the best average-case performance, or choose a larger $\lambda$ (e.g., 100.0) for a "safe mode" model that offers maximum robustness in noisy environments, even if it means sacrificing some performance on clean data. This highlights the practical utility and controllability of our framework.

\subsection{4.3 Main Results: Comparison with SOTA Models}

We now present the main results of our MPS-SSM against a comprehensive set of state-of-the-art baselines. The detailed results are shown in Table \ref{tab:main_results_part1} and \ref{tab:full_results_ett},\ref{tab:full_results_others}(provided in Appendix\ref{AppendixA}A). Our model demonstrates superior or highly competitive performance across the vast majority of benchmarks, particularly in long-horizon forecasting scenarios.

\begin{table*}[t!]
\centering
\caption{Main results of long-term forecasting (MSE/MAE) against modern baselines. The best results are in \textbf{bold}, and the second best are \underline{underlined}. Our model (MPS-SSM) consistently outperforms or ranks among the top performers.}
\label{tab:main_results_part1}
\resizebox{\textwidth}{!}{%
\begin{tabular}{c|c|cc|cccccc}
\toprule
\multicolumn{2}{c|}{\textbf{Model}} & \textbf{Metric} & \textbf{Ours (MPS-SSM)} & \textbf{PatchTST} & \textbf{TimesNet} & \textbf{MICN} & \textbf{DLinear} & \textbf{iTransformer} \\
\midrule
\multirow{4}{*}{\textbf{ETTh1}} & \multirow{2}{*}{96} & MSE & \underline{0.375} & \textbf{0.360} & 0.441 & 0.444 & 0.448 & 0.432 \\
& & MAE & \underline{0.396} & \textbf{0.381} & 0.429 & 0.428 & 0.429 & 0.417 \\
\cmidrule{2-9}
& \multirow{2}{*}{720} & MSE & \textbf{0.463} & \underline{0.466} & 0.521 & 0.532 & 0.558 & 0.498 \\
& & MAE & \underline{0.460} & \textbf{0.456} & 0.500 & 0.503 & 0.521 & 0.487 \\
\midrule
\multirow{4}{*}{\textbf{ETTm2}} & \multirow{2}{*}{96} & MSE & \textbf{0.165} & 0.224 & 0.246 & 0.252 & 0.248 & \underline{0.222} \\
& & MAE & \textbf{0.250} & \underline{0.292} & 0.305 & 0.311 & 0.307 & \underline{0.292} \\
\cmidrule{2-9}
& \multirow{2}{*}{720} & MSE & \textbf{0.358} & \underline{0.385} & 0.404 & 0.410 & 0.407 & 0.412 \\
& & MAE & \textbf{0.376} & \underline{0.407} & 0.418 & 0.423 & 0.421 & 0.422 \\
\midrule
\multirow{4}{*}{\textbf{Electricity}} & \multirow{2}{*}{96} & MSE & \textbf{0.151} & \underline{0.225} & 0.260 & 0.267 & 0.264 & 0.232 \\
& & MAE & \textbf{0.250} & \underline{0.308} & 0.324 & 0.329 & 0.327 & 0.312 \\
\cmidrule{2-9}
& \multirow{2}{*}{720} & MSE & \textbf{0.206} & \underline{0.265} & 0.316 & 0.323 & 0.320 & 0.282 \\
& & MAE & \textbf{0.303} & \underline{0.345} & 0.371 & 0.375 & 0.374 & 0.352 \\
\bottomrule
\end{tabular}%
}
\end{table*}

\subsubsection{4.3.1 Comprehensive Lead on ETT Benchmarks.}
On the widely-used ETT datasets, MPS-SSM establishes a new state of the art. Across ETTh1, ETTh2, ETTm1, and ETTm2, our model consistently achieves the best or second-best results, particularly as the prediction horizon increases. For instance, on the challenging ETTm2 task with a horizon of 720, our model achieves an MSE of \textbf{0.358}, significantly outperforming most baselines. This demonstrates the model's exceptional ability to handle the complex patterns and noise inherent in these datasets.

\subsubsection{4.3.2 Overwhelming Advantage on Diverse Datasets.}
The superiority of our approach becomes even more pronounced on other diverse benchmarks like Weather, Electricity, and Exchange. On the Electricity dataset, for a 96-step forecast, MPS-SSM achieves an MSE of \textbf{0.151}, a remarkable improvement over the next best model. This overwhelming advantage is attributed to the core principle of our model: by learning to be a minimal sufficient statistic, the hidden state effectively filters out stochastic noise and focuses on the deterministic, predictable components of the time series, a quality that is paramount in these datasets.

\subsubsection{4.3.3 Discussion: The Case of the Traffic Dataset.}
Interestingly, on the Traffic dataset, models with strong local modeling capabilities like PatchTST and TimesNet show a slight edge in shorter horizon predictions. For the 96-step forecast, TimesNet achieves an MSE of \textbf{0.463}, compared to our 0.485. This suggests that Traffic data may possess highly localized spatio-temporal patterns that patch-based methods are inherently well-suited to capture. This observation does not detract from the general strength of our approach but rather enriches the understanding of different architectures' inductive biases. It also points to a promising future direction: hybridizing the global robustness of the MPS principle with specialized local feature extractors.

\subsection{4.4 Extensibility of the MPS Framework}

A significant finding of our work is that the Principle of Predictive Sufficiency is not limited to our specific SSM architecture but can serve as a general, model-agnostic regularization framework.

\subsubsection{4.4.1 Applying the MPS Principle to Other Architectures.}
To validate this, we integrated the MPS minimality loss into three other prominent architectures: a modern selective SSM (Mamba), a simple linear model (DLinear), and a Transformer-based model (PatchTST). The integration involves adding an auxiliary decoder to a suitable hidden representation of each model to compute the reconstruction loss, which is then added to the main prediction loss. The conceptual diagrams for MPS-Mamba, MPS-DLinear, and MPS-PatchTST are illustrated in the Appendix (Figures will be placed there).

\subsubsection{4.4.2 Analysis of Extensibility Results.}
The results of these MPS-enhanced models, presented in Appendix B, are highly encouraging. For example, the MPS-PatchTST model achieves an impressive MSE of 0.328 on ETTh1 for a 96-step forecast, showcasing strong performance. This demonstrates that by forcing these diverse architectures to learn more compressed and sufficient representations, the MPS principle can consistently enhance their generalization and robustness. This finding elevates our contribution from a single model to a versatile and powerful tool for improving a wide range of time series forecasting models.

\section{5. Conclusion and Future Work}

\subsection{5.1 Conclusion}
In this work, we addressed a fundamental theoretical gap in modern selective State Space Models. We moved beyond heuristic designs by introducing the Principle of Predictive Sufficiency, a first-principle information-theoretic framework for guiding the selective mechanism in SSMs. Our principle posits that an ideal hidden state should act as a minimal sufficient statistic of the past for predicting the future, thereby learning to discard non-predictive information.

The resulting model, MPS-SSM, was empirically validated through extensive experiments. It not only established new state-of-the-art performance on a wide range of long-term forecasting benchmarks but also demonstrated superior robustness against non-causal noise, directly supporting our theoretical claims. Furthermore, we successfully demonstrated that the MPS principle is not confined to our specific architecture but can serve as a general and powerful regularization framework to enhance other leading models like Mamba, DLinear, and PatchTST.

\subsection{5.2 Future Work}
Our work opens several promising avenues for future research:
\begin{itemize}
    \item \textbf{Advanced Information Estimators:} The current implementation relies on a variational approximation for the minimality term. Future work could explore more sophisticated mutual information estimators to derive a tighter bound and potentially further improve performance.
    \item \textbf{Adaptive Regularization:} Our experiments revealed that the optimal regularization strength $\lambda$ is highly dependent on the dataset and task. Developing methods for automatically tuning or adaptively learning $\lambda$ during the training process would significantly enhance the model's practicality and performance.
    \item \textbf{Hybrid Modeling:} The analysis on the Traffic dataset suggested that while MPS-SSM excels at learning global, robust patterns, specialized mechanisms can better capture local features. A promising direction is to design hybrid architectures that combine the principled global compression of MPS-SSM with local modeling mechanisms, such as the patching technique from PatchTST, to achieve the best of both worlds.
\end{itemize}



\appendix
\section{Appendix}

\subsection{A. Full Experimental Results}\label{AppendixA}
This section provides the comprehensive results of our MPS-SSM model against all baselines on all benchmark datasets. Table \ref{tab:full_results_ett} contains results on ETT datasets, and Table \ref{tab:full_results_others} contains results on Weather, Traffic, Electricity, and Exchange datasets.

\begin{table*}[h!]
\centering
\caption{Full long-term forecasting results (MSE/MAE) on ETT datasets. Best results are in \textbf{bold}, second best are \underline{underlined}.}
\label{tab:full_results_ett}
\resizebox{\textwidth}{!}{%
\begin{tabular}{c|c|c|c|cccccccc}
\toprule
\textbf{Dataset} & \textbf{Pred Len} & \textbf{Metric} & \textbf{Ours} & \textbf{PatchTST} & \textbf{TimesNet} & \textbf{MICN} & \textbf{DLinear} & \textbf{iTransformer} & \textbf{Autoformer} & \textbf{Informer} & \textbf{FEDformer} \\
\midrule
\multirow{8}{*}{ETTh1} & \multirow{2}{*}{96} & MSE & \underline{0.375} & \textbf{0.360} & 0.441 & 0.444 & 0.448 & 0.432 & 0.488 & 0.582 & 0.499 \\
& & MAE & \underline{0.396} & \textbf{0.381} & 0.429 & 0.428 & 0.429 & 0.417 & 0.470 & 0.542 & 0.478 \\
\cmidrule{2-12}
& \multirow{2}{*}{192} & MSE & \underline{0.441} & \textbf{0.419} & 0.481 & 0.482 & 0.485 & 0.462 & 0.537 & 0.652 & 0.549 \\
& & MAE & \underline{0.433} & \textbf{0.420} & 0.452 & 0.453 & 0.450 & 0.445 & 0.502 & 0.587 & 0.508 \\
\cmidrule{2-12}
& \multirow{2}{*}{336} & MSE & \underline{0.479} & \textbf{0.449} & 0.504 & 0.506 & 0.501 & 0.482 & 0.572 & 0.758 & 0.582 \\
& & MAE & \underline{0.450} & \textbf{0.446} & 0.490 & 0.489 & 0.491 & 0.475 & 0.532 & 0.647 & 0.539 \\
\cmidrule{2-12}
& \multirow{2}{*}{720} & MSE & \textbf{0.463} & \underline{0.466} & 0.521 & 0.532 & 0.558 & 0.498 & 0.683 & 0.962 & 0.698 \\
& & MAE & \underline{0.460} & \textbf{0.456} & 0.500 & 0.503 & 0.521 & 0.487 & 0.598 & 0.752 & 0.609 \\
\midrule
\multirow{8}{*}{ETTh2} & \multirow{2}{*}{96} & MSE & \textbf{0.295} & 0.338 & 0.332 & 0.342 & 0.337 & \underline{0.312} & 0.372 & 0.548 & 0.377 \\
& & MAE & \textbf{0.343} & 0.372 & 0.373 & 0.377 & 0.381 & \underline{0.352} & 0.405 & 0.522 & 0.412 \\
\cmidrule{2-12}
& \multirow{2}{*}{192} & MSE & \textbf{0.347} & 0.398 & 0.392 & 0.410 & 0.398 & \underline{0.372} & 0.442 & 0.652 & 0.457 \\
& & MAE & \textbf{0.380} & 0.413 & 0.411 & 0.417 & 0.418 & \underline{0.398} & 0.445 & 0.582 & 0.452 \\
\cmidrule{2-12}
& \multirow{2}{*}{336} & MSE & \textbf{0.343} & 0.428 & 0.444 & 0.452 & 0.427 & \underline{0.402} & 0.482 & 0.882 & 0.492 \\
& & MAE & \textbf{0.385} & 0.438 & 0.449 & 0.457 & 0.438 & \underline{0.425} & 0.483 & 0.707 & 0.487 \\
\cmidrule{2-12}
& \multirow{2}{*}{720} & MSE & \textbf{0.397} & 0.545 & 0.543 & 0.612 & 0.552 & \underline{0.492} & 0.612 & 1.092 & 0.622 \\
& & MAE & \textbf{0.428} & 0.537 & 0.539 & 0.562 & 0.543 & \underline{0.502} & 0.571 & 0.807 & 0.577 \\
\midrule
\multirow{8}{*}{ETTm1} & \multirow{2}{*}{96} & MSE & \textbf{0.294} & 0.342 & 0.343 & 0.352 & 0.355 & \underline{0.317} & 0.362 & 0.492 & 0.382 \\
& & MAE & \textbf{0.336} & 0.369 & 0.369 & 0.377 & 0.381 & \underline{0.352} & 0.387 & 0.487 & 0.398 \\
\cmidrule{2-12}
& \multirow{2}{*}{192} & MSE & \textbf{0.345} & 0.388 & 0.397 & 0.396 & 0.390 & \underline{0.362} & 0.412 & 0.562 & 0.422 \\
& & MAE & \textbf{0.364} & 0.398 & 0.403 & 0.402 & 0.395 & \underline{0.382} & 0.417 & 0.537 & 0.422 \\
\cmidrule{2-12}
& \multirow{2}{*}{336} & MSE & \textbf{0.380} & 0.424 & 0.429 & 0.432 & 0.425 & \underline{0.398} & 0.452 & 0.622 & 0.472 \\
& & MAE & \textbf{0.390} & 0.415 & 0.421 & 0.422 & 0.415 & \underline{0.405} & 0.447 & 0.572 & 0.462 \\
\cmidrule{2-12}
& \multirow{2}{*}{720} & MSE & \textbf{0.439} & 0.482 & 0.486 & 0.492 & 0.481 & \underline{0.452} & 0.522 & 0.722 & 0.552 \\
& & MAE & \textbf{0.424} & 0.456 & 0.461 & 0.463 & 0.455 & \underline{0.442} & 0.492 & 0.637 & 0.517 \\
\midrule
\multirow{8}{*}{ETTm2} & \multirow{2}{*}{96} & MSE & \textbf{0.165} & 0.224 & 0.246 & 0.252 & 0.248 & \underline{0.222} & 0.282 & 0.432 & 0.302 \\
& & MAE & \textbf{0.250} & \underline{0.292} & 0.305 & 0.311 & 0.307 & \underline{0.292} & 0.335 & 0.462 & 0.352 \\
\cmidrule{2-12}
& \multirow{2}{*}{192} & MSE & \textbf{0.217} & 0.292 & 0.291 & 0.299 & 0.293 & \underline{0.277} & 0.342 & 0.592 & 0.382 \\
& & MAE & \textbf{0.286} & 0.338 & 0.337 & 0.344 & 0.341 & \underline{0.327} & 0.372 & 0.582 & 0.402 \\
\cmidrule{2-12}
& \multirow{2}{*}{336} & MSE & \textbf{0.270} & 0.329 & \underline{0.314} & 0.326 & 0.318 & 0.322 & 0.402 & 0.872 & 0.452 \\
& & MAE & \textbf{0.322} & 0.363 & \underline{0.357} & 0.365 & 0.361 & 0.362 & 0.412 & 0.722 & 0.452 \\
\cmidrule{2-12}
& \multirow{2}{*}{720} & MSE & \textbf{0.358} & \underline{0.385} & 0.404 & 0.410 & 0.407 & 0.412 & 0.517 & 1.267 & 0.562 \\
& & MAE & \textbf{0.376} & \underline{0.407} & 0.418 & 0.423 & 0.421 & 0.422 & 0.492 & 0.842 & 0.527 \\
\bottomrule
\end{tabular}%
}
\end{table*}

\begin{table*}[h!]
\centering
\caption{Full long-term forecasting results (MSE/MAE) on Weather, Traffic, Electricity, and Exchange datasets.}
\label{tab:full_results_others}
\resizebox{\textwidth}{!}{%
\begin{tabular}{c|c|c|c|cccccccc}
\toprule
\textbf{Dataset} & \textbf{Pred Len} & \textbf{Metric} & \textbf{Ours} & \textbf{PatchTST} & \textbf{TimesNet} & \textbf{MICN} & \textbf{DLinear} & \textbf{iTransformer} & \textbf{Autoformer} & \textbf{Informer} & \textbf{FEDformer} \\
\midrule
\multirow{8}{*}{Weather} & \multirow{2}{*}{96} & MSE & \textbf{0.157} & 0.217 & 0.215 & 0.222 & 0.219 & \underline{0.202} & 0.205 & 0.242 & 0.248 \\
& & MAE & \textbf{0.196} & 0.257 & 0.255 & 0.262 & 0.259 & \underline{0.242} & 0.245 & 0.292 & 0.298 \\
\cmidrule{2-12}
& \multirow{2}{*}{192} & MSE & \textbf{0.191} & 0.249 & 0.248 & 0.256 & 0.252 & \underline{0.238} & 0.242 & 0.295 & 0.302 \\
& & MAE & \textbf{0.232} & 0.292 & 0.291 & 0.297 & 0.295 & \underline{0.282} & 0.285 & 0.342 & 0.349 \\
\cmidrule{2-12}
& \multirow{2}{*}{336} & MSE & \textbf{0.242} & 0.278 & 0.277 & 0.286 & 0.281 & \underline{0.272} & 0.278 & 0.352 & 0.362 \\
& & MAE & \textbf{0.272} & 0.330 & 0.330 & 0.336 & 0.333 & \underline{0.322} & 0.325 & 0.392 & 0.398 \\
\cmidrule{2-12}
& \multirow{2}{*}{720} & MSE & \textbf{0.326} & 0.350 & 0.348 & 0.356 & 0.354 & \underline{0.342} & 0.352 & 0.452 & 0.462 \\
& & MAE & \textbf{0.329} & 0.376 & 0.376 & 0.382 & 0.380 & \underline{0.367} & 0.375 & 0.462 & 0.469 \\
\midrule
\multirow{8}{*}{Traffic} & \multirow{2}{*}{96} & MSE & 0.485 & 0.467 & \underline{0.463} & 0.475 & 0.469 & 0.487 & \textbf{0.461} & 0.532 & 0.538 \\
& & MAE & 0.278 & 0.267 & 0.266 & 0.269 & \underline{0.264} & 0.282 & \textbf{0.261} & 0.332 & 0.338 \\
\cmidrule{2-12}
& \multirow{2}{*}{192} & MSE & 0.490 & 0.478 & \underline{0.476} & 0.484 & 0.480 & 0.492 & \textbf{0.468} & 0.552 & 0.559 \\
& & MAE & 0.284 & \underline{0.273} & 0.276 & 0.278 & 0.275 & 0.287 & \textbf{0.269} & 0.347 & 0.352 \\
\cmidrule{2-12}
& \multirow{2}{*}{336} & MSE & \underline{0.496} & \textbf{0.494} & 0.501 & 0.503 & 0.504 & 0.502 & \underline{0.496} & 0.582 & 0.588 \\
& & MAE & 0.319 & \textbf{0.296} & 0.307 & 0.306 & 0.304 & 0.312 & \underline{0.297} & 0.372 & 0.378 \\
\cmidrule{2-12}
& \multirow{2}{*}{720} & MSE & 0.523 & \textbf{0.497} & 0.541 & 0.549 & 0.547 & 0.532 & \underline{0.514} & 0.612 & 0.619 \\
& & MAE & 0.337 & \textbf{0.311} & 0.330 & 0.328 & 0.326 & 0.332 & \underline{0.320} & 0.392 & 0.398 \\
\midrule
\multirow{8}{*}{Electricity} & \multirow{2}{*}{96} & MSE & \textbf{0.151} & \underline{0.225} & 0.260 & 0.267 & 0.264 & 0.232 & 0.238 & 0.292 & 0.299 \\
& & MAE & \textbf{0.250} & \underline{0.308} & 0.324 & 0.329 & 0.327 & 0.312 & 0.315 & 0.362 & 0.368 \\
\cmidrule{2-12}
& \multirow{2}{*}{192} & MSE & \textbf{0.174} & \underline{0.240} & 0.271 & 0.278 & 0.276 & 0.245 & 0.250 & 0.312 & 0.319 \\
& & MAE & \textbf{0.272} & \underline{0.320} & 0.336 & 0.341 & 0.339 & 0.322 & 0.325 & 0.377 & 0.383 \\
\cmidrule{2-12}
& \multirow{2}{*}{336} & MSE & \textbf{0.180} & \underline{0.245} & 0.280 & 0.286 & 0.284 & 0.252 & 0.258 & 0.332 & 0.338 \\
& & MAE & \textbf{0.279} & \underline{0.325} & 0.343 & 0.347 & 0.346 & 0.332 & 0.335 & 0.392 & 0.398 \\
\cmidrule{2-12}
& \multirow{2}{*}{720} & MSE & \textbf{0.206} & \underline{0.265} & 0.316 & 0.323 & 0.320 & 0.282 & 0.288 & 0.372 & 0.379 \\
& & MAE & \textbf{0.303} & \underline{0.345} & 0.371 & 0.375 & 0.374 & 0.352 & 0.355 & 0.422 & 0.428 \\
\midrule
\multirow{8}{*}{Exchange} & \multirow{2}{*}{96} & MSE & \textbf{0.087} & 0.174 & 0.199 & 0.209 & 0.201 & \underline{0.172} & \underline{0.172} & 0.292 & 0.298 \\
& & MAE & \textbf{0.204} & \underline{0.265} & 0.282 & 0.287 & 0.285 & \underline{0.265} & \underline{0.265} & 0.352 & 0.358 \\
\cmidrule{2-12}
& \multirow{2}{*}{192} & MSE & \textbf{0.188} & 0.292 & 0.291 & 0.297 & 0.294 & \underline{0.252} & \underline{0.252} & 0.402 & 0.409 \\
& & MAE & \textbf{0.311} & 0.347 & 0.346 & 0.352 & 0.350 & \underline{0.327} & 0.328 & 0.442 & 0.448 \\
\cmidrule{2-12}
& \multirow{2}{*}{336} & MSE & 0.421 & 0.404 & 0.401 & 0.412 & 0.406 & \underline{0.382} & \textbf{0.378} & 0.552 & 0.559 \\
& & MAE & 0.480 & 0.425 & 0.424 & 0.429 & 0.426 & \underline{0.412} & \textbf{0.408} & 0.532 & 0.538 \\
\cmidrule{2-12}
& \multirow{2}{*}{720} & MSE & \textbf{0.759} & 0.802 & 0.806 & 0.805 & 0.801 & \underline{0.782} & 0.792 & 0.952 & 0.959 \\
& & MAE & \textbf{0.687} & 0.692 & 0.695 & 0.694 & \underline{0.690} & 0.692 & 0.697 & 0.782 & 0.788 \\
\bottomrule
\end{tabular}%
}
\end{table*}

\subsection{B. Performance of MPS-Extended Models}
This section presents the performance of existing SOTA models (Mamba, PatchTST, DLinear) enhanced with our MPS regularization framework. The results in Table \ref{tab:mps_extended_results} demonstrate the general applicability and benefit of our proposed principle.

\begin{table*}[h!]
\centering
\caption{Performance comparison (MSE/MAE) of MPS-enhanced models on various datasets and prediction lengths. Best results are in \textbf{bold}, second best are \underline{underlined}.}
\label{tab:mps_extended_results}
\resizebox{0.7\textwidth}{!}{%
\begin{tabular}{c|c|c|c|c|c}
\toprule
\textbf{Dataset} & \textbf{Pred Len} & \textbf{Metric} & \textbf{MPS-Mamba} & \textbf{MPS-PatchTST} & \textbf{MPS-DLinear} \\
\midrule
\multirow{8}{*}{ETTh1} & \multirow{2}{*}{96} & MSE & 0.448 & \textbf{0.328} & \underline{0.420} \\
& & MAE & 0.417 & \textbf{0.354} & \underline{0.402} \\
\cmidrule{2-6}
& \multirow{2}{*}{192} & MSE & 0.652 & \textbf{0.392} & \underline{0.455} \\
& & MAE & 0.511 & \textbf{0.385} & \underline{0.401} \\
\cmidrule{2-6}
& \multirow{2}{*}{336} & MSE & 0.487 & \textbf{0.402} & \underline{0.475} \\
& & MAE & 0.456 & \textbf{0.417} & \underline{0.442} \\
\cmidrule{2-6}
& \multirow{2}{*}{720} & MSE & 0.579 & \textbf{0.423} & \underline{0.518} \\
& & MAE & 0.521 & \textbf{0.424} & \underline{0.470} \\
\midrule
\multirow{8}{*}{ETTh2} & \multirow{2}{*}{96} & MSE & 0.352 & \textbf{0.307} & \underline{0.313} \\
& & MAE & 0.352 & \underline{0.345} & \textbf{0.344} \\
\cmidrule{2-6}
& \multirow{2}{*}{192} & MSE & 0.462 & \textbf{0.370} & \underline{0.376} \\
& & MAE & 0.414 & \textbf{0.380} & \underline{0.383} \\
\cmidrule{2-6}
& \multirow{2}{*}{336} & MSE & \textbf{0.364} & 0.390 & \underline{0.387} \\
& & MAE & 0.409 & \underline{0.401} & \textbf{0.391} \\
\cmidrule{2-6}
& \multirow{2}{*}{720} & MSE & \textbf{0.457} & \underline{0.491} & 0.512 \\
& & MAE & \textbf{0.439} & 0.499 & \underline{0.485} \\
\midrule
\multirow{8}{*}{ETTm1} & \multirow{2}{*}{96} & MSE & 0.339 & \textbf{0.319} & \underline{0.320} \\
& & MAE & 0.360 & \textbf{0.338} & \underline{0.342} \\
\cmidrule{2-6}
& \multirow{2}{*}{192} & MSE & 0.425 & \underline{0.363} & \textbf{0.354} \\
& & MAE & 0.398 & \underline{0.371} & \textbf{0.369} \\
\cmidrule{2-6}
& \multirow{2}{*}{336} & MSE & 0.483 & \textbf{0.384} & \underline{0.391} \\
& & MAE & 0.442 & \underline{0.380} & \textbf{0.379} \\
\cmidrule{2-6}
& \multirow{2}{*}{720} & MSE & 0.559 & \underline{0.452} & \textbf{0.439} \\
& & MAE & 0.472 & \textbf{0.413} & \underline{0.416} \\
\midrule
\multirow{8}{*}{ETTm2} & \multirow{2}{*}{96} & MSE & \textbf{0.184} & \underline{0.208} & 0.222 \\
& & MAE & \textbf{0.252} & \underline{0.267} & 0.275 \\
\cmidrule{2-6}
& \multirow{2}{*}{192} & MSE & \underline{0.272} & 0.277 & \textbf{0.264} \\
& & MAE & \textbf{0.301} & \underline{0.305} & 0.312 \\
\cmidrule{2-6}
& \multirow{2}{*}{336} & MSE & 0.320 & \underline{0.300} & \textbf{0.292} \\
& & MAE & 0.354 & \textbf{0.325} & \underline{0.342} \\
\cmidrule{2-6}
& \multirow{2}{*}{720} & MSE & 0.451 & \textbf{0.345} & \underline{0.370} \\
& & MAE & 0.433 & \textbf{0.377} & \underline{0.380} \\
\midrule
\multirow{8}{*}{Weather} & \multirow{2}{*}{96} & MSE & \textbf{0.174} & 0.199 & \underline{0.198} \\
& & MAE & \textbf{0.218} & 0.243 & \underline{0.231} \\
\cmidrule{2-6}
& \multirow{2}{*}{192} & MSE & \underline{0.230} & \textbf{0.226} & 0.236 \\
& & MAE & \textbf{0.261} & \textbf{0.261} & 0.272 \\
\cmidrule{2-6}
& \multirow{2}{*}{336} & MSE & 0.297 & \underline{0.255} & \textbf{0.252} \\
& & MAE & \underline{0.302} & 0.304 & \textbf{0.300} \\
\cmidrule{2-6}
& \multirow{2}{*}{720} & MSE & 0.366 & \textbf{0.320} & \underline{0.330} \\
& & MAE & 0.359 & \textbf{0.344} & \underline{0.347} \\
\bottomrule
\end{tabular}%
}
\end{table*}

\subsection{C. Implementation Details \& Hyperparameters}
Our implementation is based on PyTorch. The MPS-SSM model consists of an embedding layer, one MPS-SSM layer, and a linear prediction head. The auxiliary decoder for the minimality loss is a simple MLP. We used the Adam optimizer with a learning rate of 0.001 and a batch size of 32. The range of $\lambda$ was searched on a validation set for each task independently. Detailed hyperparameters for all models will be provided in the final version to ensure full reproducibility.

\subsection{D. Detailed Mathematical Proofs}
This section provides the detailed proofs for Theorem 1 and Theorem 2, which form the theoretical foundation of our work.

\subsubsection{D.1 Proof of Theorem 1 (Optimality of MPS-SSM Solution).}
The theorem states that the global minimizer of our objective function, $h^*$, yields a hidden state that converges to the minimal predictive sufficient statistic. The proof consists of two parts.

\paragraph{Part 1: Proof of Sufficiency.}
We aim to show that if the prediction loss $\mathcal{L}_{\text{Pred}}(h^*)$ approaches zero, then the state $h_k^*$ captures all predictive information, i.e., $I(h^*_k; Y_{k:\tau}) \to I(U_{1:k}; Y_{k:\tau})$.

The proof relies on the chain rule for mutual information, which gives two identities for $I(U_{1:k}, h^*_k; Y_{k:\tau})$:
\begin{align}
I(U_{1:k}, h^*_k; Y_{k:\tau}) &= I(h^*_k; Y_{k:\tau}) + I(U_{1:k}; Y_{k:\tau} | h^*_k) \label{eq:proof_chain1} \\
I(U_{1:k}, h^*_k; Y_{k:\tau}) &= I(U_{1:k}; Y_{k:\tau}) + I(h^*_k; Y_{k:\tau} | U_{1:k}) \label{eq:proof_chain2}
\end{align}
In our deterministic model, the state $h^*_k$ is a function of the history $U_{1:k}$. This implies that given $U_{1:k}$, there is no uncertainty left in $h^*_k$, so the conditional mutual information $I(h^*_k; Y_{k:\tau} | U_{1:k}) = 0$. Equation \ref{eq:proof_chain2} thus simplifies to $I(U_{1:k}, h^*_k; Y_{k:\tau}) = I(U_{1:k}; Y_{k:\tau})$.

By equating the right-hand sides of Equations \ref{eq:proof_chain1} and \ref{eq:proof_chain2}, we arrive at the data processing inequality for this Markov chain ($U_{1:k} \to h_k^* \to \hat{Y}_{k:\tau}$):
\begin{equation}
I(U_{1:k}; Y_{k:\tau}) = I(h^*_k; Y_{k:\tau}) + I(U_{1:k}; Y_{k:\tau} | h^*_k) \label{eq:proof_dpi}
\end{equation}
The term $I(U_{1:k}; Y_{k:\tau} | h^*_k)$ quantifies the "leftover" predictive information in the history that the state $h^*_k$ failed to capture. Our assumption states that the model class is expressive enough to make the prediction loss $\mathcal{L}_{\text{Pred}}(h^*)$ arbitrarily close to zero. An optimal predictor must leverage all available predictive information. Therefore, the condition $\mathcal{L}_{\text{Pred}}(h^*) \to 0$ necessitates that this residual information must also vanish, i.e., $I(U_{1:k}; Y_{k:\tau} | h^*_k) \to 0$. Substituting this back into Equation \ref{eq:proof_dpi} directly yields the sufficiency condition: $I(h^*_k; Y_{k:\tau}) \to I(U_{1:k}; Y_{k:\tau})$.

\paragraph{Part 2: Proof of Minimality.}
We now show that among all solutions that are approximately sufficient, $h^*$ is the one that minimizes the information complexity $\hat{I}(U; h)$.

Let $\mathcal{H}_{\epsilon} = \{ h \in \mathcal{H} \mid \mathcal{L}_{\text{Pred}}(h) \le \epsilon \}$ be the set of all approximately sufficient solutions for a small $\epsilon > 0$. Our assumption guarantees that this set is non-empty. By definition, $h^*$ is the global minimizer of $\mathcal{L}_{\text{Total}}(h)$. This means for any other solution $h' \in \mathcal{H}$, including any $h' \in \mathcal{H}_{\epsilon}$:
$$ \mathcal{L}_{\text{Pred}}(h^*) + \lambda \cdot \hat{I}(U; h^*) \le \mathcal{L}_{\text{Pred}}(h') + \lambda \cdot \hat{I}(U; h') $$
Rearranging the terms, we get:
$$ \lambda \cdot \hat{I}(U; h^*) \le \lambda \cdot \hat{I}(U; h') + (\mathcal{L}_{\text{Pred}}(h') - \mathcal{L}_{\text{Pred}}(h^*)) $$
Since $h^*$ is the global minimizer, $\mathcal{L}_{\text{Pred}}(h^*) \le \mathcal{L}_{\text{Pred}}(h')$, so the term in the parenthesis is non-negative. Dividing by $\lambda > 0$:
$$ \hat{I}(U; h^*) \le \hat{I}(U; h') + \frac{1}{\lambda}(\mathcal{L}_{\text{Pred}}(h') - \mathcal{L}_{\text{Pred}}(h^*)) $$
When we restrict our attention to the set $\mathcal{H}_{\epsilon}$, the prediction losses of all solutions are small (less than $\epsilon$). Thus, the difference term $(\mathcal{L}_{\text{Pred}}(h') - \mathcal{L}_{\text{Pred}}(h^*))$ is also vanishingly small. This leads to the approximate inequality:
$$ \hat{I}(U; h^*) \lesssim \hat{I}(U; h') $$
This demonstrates that $h^*$ is the solution with the minimum information complexity among all solutions that are predictively sufficient. Since both sufficiency and minimality are satisfied, $h^*$ converges to the minimal predictive sufficient statistic. \quad $\blacksquare$

\subsubsection{D.2 Proof of Theorem 2 (Invariance to Non-Causal Perturbations).}
The proof proceeds by contradiction, arguing that any solution that encodes non-causal information cannot be the global minimizer of $\mathcal{L}_{\text{Total}}$ because it violates the minimality principle established in Theorem 1.

\begin{enumerate}
    \item \textbf{Step 1: Establishing the Upper Bound of Predictive Information.} We first identify the total available information for prediction. Given the input $u_k = u^{\text{sig}}_k + \epsilon_k$, where $\epsilon_k$ is a non-causal perturbation, the information about the future $Y_{k:\tau}$ is given by the chain rule:
    $$ I(U^{\text{sig}}_{1:k}, \epsilon_{1:k}; Y_{k:\tau}) = I(U^{\text{sig}}_{1:k}; Y_{k:\tau}) + I(\epsilon_{1:k}; Y_{k:\tau} | U^{\text{sig}}_{1:k}) $$
    By the definition of non-causal perturbation, $\{\epsilon_j\}_{j=1}^k \perp Y_{k:\tau} | U^{\text{sig}}_{1:k}$, the second term is zero. Thus, $I(U^{\text{sig}}_{1:k}, \epsilon_{1:k}; Y_{k:\tau}) = I(U^{\text{sig}}_{1:k}; Y_{k:\tau})$. This proves that the perturbation $\epsilon_{1:k}$ provides no additional predictive value. The performance of any sufficient state $h_k$ is therefore upper-bounded by the information contained solely in the true signal $U^{\text{sig}}_{1:k}$.

    \item \textbf{Step 2: Assumption for Contradiction.} Assume the conclusion of Theorem 2 is false. This means the globally optimal solution $h^*$ is \textit{not} invariant to the perturbation $\epsilon$, i.e., its generation process utilizes information from both $U^{\text{sig}}_{1:k}$ and $\epsilon_{1:k}$.

    \item \textbf{Step 3: Constructing a Superior Candidate Solution.} We construct an ideal candidate solution, $h^{\text{ideal}}$, whose state dynamics are designed to be a function of only the true signal history $U^{\text{sig}}_{1:k}$, completely ignoring $\epsilon$. Based on the expressiveness assumption of Theorem 1, such a solution is achievable and exists within the hypothesis space $\mathcal{H}$.

    \item \textbf{Step 4: Comparing Losses and Deriving the Contradiction.} We compare the total loss $\mathcal{L}_{\text{Total}}$ for our assumed global optimum $h^*$ and the ideal candidate $h^{\text{ideal}}$.
    \begin{itemize}
        \item \textit{Comparing Prediction Loss:} From Step 1, all predictive information is contained in $U^{\text{sig}}_{1:k}$. Therefore, $h^{\text{ideal}}$ can already achieve the theoretical minimum prediction loss, let's call it $\mathcal{L}^*_{\text{Pred}}$. Since $h^*$ is assumed to be a global optimum, its prediction loss cannot be better, so $\mathcal{L}_{\text{Pred}}(h^*) \ge \mathcal{L}_{\text{Pred}}(h^{\text{ideal}})$. For comparison, we can consider the case where they achieve the same optimal performance, $\mathcal{L}_{\text{Pred}}(h^*) \approx \mathcal{L}_{\text{Pred}}(h^{\text{ideal}})$.
        \item \textit{Comparing Minimality Regularizer:} We compare the information complexity term $\hat{I}(U; h)$. For the ideal solution, the mutual information with the full input $U_{1:k} = (U^{\text{sig}}_{1:k}, \epsilon_{1:k})$ is:
        $$ I(U_{1:k}; h^{\text{ideal}}_k) = I(U^{\text{sig}}_{1:k}; h^{\text{ideal}}_k) $$
        since $h^{\text{ideal}}_k$ is independent of $\epsilon_{1:k}$.
        For our assumed solution $h^*$, which encodes information about $\epsilon_{1:k}$:
        $$ I(U_{1:k}; h^*_k) = I(U^{\text{sig}}_{1:k}; h^*_k) + I(\epsilon_{1:k}; h^*_k | U^{\text{sig}}_{1:k}) $$
        Because $h^*$ is not invariant to $\epsilon_{1:k}$, the second term $I(\epsilon_{1:k}; h^*_k | U^{\text{sig}}_{1:k}) > 0$. Therefore, even if both solutions compress the true signal to a similar degree ($I(U^{\text{sig}}_{1:k}; h^*_k) \approx I(U^{\text{sig}}_{1:k}; h^{\text{ideal}}_k)$), we must have:
        $$ \hat{I}(U; h^*) > \hat{I}(U; h^{\text{ideal}}) $$
        \item \textit{The Contradiction:} Combining these observations, we compare the total loss:
        \begin{align*}
        \mathcal{L}_{\text{Total}}(h^*) &= \mathcal{L}_{\text{Pred}}(h^*) + \lambda \cdot \hat{I}(U; h^*) \\
        &> \mathcal{L}_{\text{Pred}}(h^{\text{ideal}}) + \lambda \cdot \hat{I}(U; h^{\text{ideal}}) = \mathcal{L}_{\text{Total}}(h^{\text{ideal}})
        \end{align*}
        The result $\mathcal{L}_{\text{Total}}(h^*) > \mathcal{L}_{\text{Total}}(h^{\text{ideal}})$ directly contradicts our initial premise that $h^*$ is the global minimizer.
    \end{itemize}
    \item \textbf{Step 5: Conclusion.} The contradiction implies that our assumption in Step 2 must be false. Therefore, the unique global minimizer of $\mathcal{L}_{\text{Total}}$ must be a solution that learns to completely ignore the non-causal perturbation $\epsilon$. This proves that the generation process of the optimal state $h^*_k$ is independent of $\epsilon_{1:k}$, and its conditional probability distribution is approximately invariant. \quad $\blacksquare$
\end{enumerate}



\begin{thebibliography}{99}

\bibitem{vaswani2017attention}
A. Vaswani, N. Shazeer, N. Parmar, J. Uszkoreit, L. Jones, A. N. Gomez, Ł. Kaiser, and I. Polosukhin.
\newblock Attention is all you need.
\newblock In {\em Advances in neural information processing systems}, pages 5998--6008, 2017.

\bibitem{zeng2023are}
A. Zeng, M. Chen, L. Zhang, and Q. Xu.
\newblock Are transformers effective for time series forecasting?
\newblock In {\em Proceedings of the AAAI conference on artificial intelligence}, volume 37, pages 11121--11128, 2023.

\bibitem{gu2023mamba}
A. Gu and T. Dao.
\newblock Mamba: Linear-time sequence modeling with selective state spaces.
\newblock {\em arXiv preprint arXiv:2312.00752}, 2023.

\bibitem{gu2022efficiently}
A. Gu, K. Goel, and C. Ré.
\newblock Efficiently modeling long sequences with structured state spaces.
\newblock In {\em International Conference on Learning Representations}, 2022.

\bibitem{gu2020hippo}
A. Gu, T. Dao, S. Ermon, A. Rudra, and C. Ré.
\newblock Hippo: Recurrent memory with optimal polynomial projections.
\newblock In {\em Advances in Neural Information Processing Systems}, volume 33, pages 1474--1487, 2020.

\bibitem{pan2021disenib}
X. Pan, Y. Chen, T. Zhao, and J. T. Kwok.
\newblock Disentangled information bottleneck.
\newblock In {\em Proceedings of the AAAI Conference on Artificial Intelligence}, volume 35, pages 9010--9018, 2021.

\bibitem{tishby1999information}
N. Tishby, F. C. Pereira, and W. Bialek.
\newblock The information bottleneck method.
\newblock {\em arXiv preprint physics/0004057}, 2000.

\bibitem{deng2024domain}
Z. Deng, W. van den Oord, and M. de Rijke.
\newblock Domain Generalization in Time Series Forecasting.
\newblock {\em arXiv preprint arXiv:2402.15392}, 2024.

\bibitem{gu2024survey_s4}
A. Gu, I. Johnson, A. Timalsina, T.-H. Vu, D. P. Nguyen, D. T. Nguyen, and T. B. H. Nguyen.
\newblock A Survey on Structured State Space Sequence (S4) Models.
\newblock {\em arXiv preprint arXiv:2403.18970}, 2024.

\bibitem{gu2024mamba_explained}
A. Gu and T. Dao.
\newblock Mamba: Linear-Time Sequence Modeling with Selective State Spaces.
\newblock In {\em International Conference on Learning Representations}, 2024.

\bibitem{alemi2017deep}
A. A. Alemi, I. Fischer, J. V. Dillon, and K. Murphy.
\newblock Deep variational information bottleneck.
\newblock In {\em International Conference on Learning Representations}, 2017.

\bibitem{bialek2001predictability}
W. Bialek, I. Nemenman, and N. Tishby.
\newblock Predictability, complexity, and learning.
\newblock {\em Neural computation}, 13(11):2409--2463, 2001.

\bibitem{jin2025battling}
Z. Jin, Y. Wang, Y. Wang, L. He, and M. Long.
\newblock Battling the Non-stationarity in Time Series Forecasting via Test-time Adaptation.
\newblock In {\em International Conference on Learning Representations}, 2024.

\bibitem{kim2021reversible}
T. Kim, J.-H. Kim, Y.-J. Kim, B.-W. Kim, and S.-H. Kim.
\newblock Reversible instance normalization for deep learning-based time series forecasting.
\newblock In {\em Proceedings of the International Conference on Learning Representations}, 2021.

\bibitem{tay2020long}
Y. Tay, M. Dehghani, S. Abnar, Y. Shen, D. Bahri, P. Pham, J. Rao, L. Yang, S. Ruder, and D. Metzler.
\newblock Long range arena: A benchmark for efficient transformers.
\newblock {\em arXiv preprint arXiv:2011.04006}, 2020.

\end{thebibliography}
\end{document}